\documentclass[conference]{IEEEtran}
\IEEEoverridecommandlockouts
\usepackage{cite}
\usepackage{amsmath,amssymb,amsfonts}
\usepackage{algorithm}
\usepackage{algpseudocode}
\usepackage{graphicx}
\usepackage{textcomp}
\usepackage{xcolor}
\usepackage[]{nohyperref} 
\usepackage{url}
\usepackage{pgfplots, tikz}
\usepgfplotslibrary{groupplots}
\usepackage{enumitem}
\usepackage{subcaption}
\usepackage{caption}
\pgfplotsset{compat=1.12}
\usepackage{booktabs}
\usepackage{tabularx}
\usepackage{float}
\usepackage{amsthm, bm, siunitx}
\DeclareSIUnit{\belmilliwatt}{Bm}
\DeclareSIUnit{\dBm}{\deci\belmilliwatt}

\theoremstyle{definition}

\newcommand{\normsq}[1]{\left\lVert#1\right\Vert_{2}^{2}}

\newcommand{\cov}[1]{\bm{C}_{#1}}

\newcommand{\gauss}[1]{\mathcal{N}\left(\bm{0}, #1\right)}

\newcommand{\tran}{^{\mathsf{T}}}

\newcommand{\C}{\mathbb{C}}
\newcommand{\R}{\mathbb{R}}
\newcommand{\ex}{\mathrm{e}}
\newcommand{\jma}{\mathrm{j}}

\newcommand{\x}{\bm{x}}
\newcommand{\y}{\bm{y}}
\newcommand{\ch}{\bm{H}}

\newcommand{\z}{\bm{z}}

\newcommand{\wa}{\bm{w}}

\newcommand{\bxi}{\bm{\xi}}

\newcommand{\btau}{\bm{\tau}}

\newcommand{\btheta}{\bm{\theta}}
\newcommand{\bpsi}{\bm{\psi}}

\newcommand{\obs}{\bm{o}}

\newcommand{\ntq}{N_{T_{q}}}
\newcommand{\nr}{N_{R}}
\newcommand{\nrv}{N_{R_{v}}}

\newcommand{\nsc}{N}

\newcommand{\symset}{\mathcal{K}}
\newcommand{\carrierset}{\mathcal{N}}
\newcommand{\resourceset}{\mathcal{R}}
\newcommand{\userset}{\mathcal{Q}}
\newcommand{\rxset}{\mathcal{V}}

\def\BibTeX{{\rm B\kern-.05em{\sc i\kern-.025em b}\kern-.08em
    T\kern-.1667em\lower.7ex\hbox{E}\kern-.125emX}}

\IEEEoverridecommandlockouts \IEEEpubid{\makebox[\columnwidth]{978-8-3503-8544-1/24/\$31.00~\copyright{}2024 IEEE \hfill} \hspace{\columnsep}\makebox[\columnwidth]{ }}

\begin{document}
\title{A Digital Twinning Platform for Integrated Sensing, Communications and Robotics
\thanks{
The authors were supported in part by the
German Federal Ministry of Education and Research (BMBF)
in the programme “Souver\"an. Digital. Vernetzt.”
within the research hub 6G-life under Grant 16KISK002,
and also by the Bavarian Ministry of Economic Affairs,
Regional Development and Energy within the project 6G Future Lab Bavaria.
U. M\"onich and H. Boche were also supported by the BMBF within the project "Post Shannon Communication - NewCom" under Grant 16KIS1003K.
M. Fees and A. Feik contributed to this work during their studies at TU Munich. 
}
}

\author{\IEEEauthorblockN{
Vlad C. Andrei\IEEEauthorrefmark{1},
Xinyang Li\IEEEauthorrefmark{1}, 
Maresa Fees\IEEEauthorrefmark{1},
Andreas Feik\IEEEauthorrefmark{2}, 
Ullrich J. M\"onich\IEEEauthorrefmark{1},
Holger Boche\IEEEauthorrefmark{1}\IEEEauthorrefmark{3}
\IEEEauthorblockA{\IEEEauthorrefmark{1}Chair of Theoretical Information Technology, Technical University of Munich, Munich, Germany\\
\IEEEauthorrefmark{1}BMBF Research Hub 6G-life,\\
\IEEEauthorrefmark{2}Department of Information Technology and Electrical Engineering, ETH Zurich, Switzerland, \\
\IEEEauthorrefmark{3}Munich Center for Quantum Science and Technology, Munich, Germany\\
Emails: \{vlad.andrei, 
xinyang.li,
maresa.fees,
moenich,
boche\}@tum.de, anfeik@ethz.ch }
}}

\maketitle

\begin{abstract}
In this paper, a digital twinning framework for indoor integrated sensing, communications, and robotics is proposed, designed, and implemented. 
Besides leveraging powerful robotics and ray-tracing technologies, the framework also enables integration with real-world sensors and reactive updates triggered by changes in the environment. 
The framework is designed with commercial, off-the-shelf components in mind, thus facilitating experimentation in the different areas of communication, sensing, and robotics. 
Experimental results showcase the feasibility and accuracy of indoor localization using digital twins and validate our implementation both qualitatively and quantitatively.
\end{abstract}


\section{Introduction} \label{sec:intro}

With the successful deployment and commercialization of 5G systems, 6G is expected to revolutionize communication technologies by offering significantly higher data speeds, reduced latency, and enhanced reliability\cite{walid6g,holger6g}.
As one of the key technologies driving the next-generation networks, integrated sensing and communications (ISAC) provides future wireless systems with the ability to ``see" the real world and facilitates the interactivity of both physical and digital environments\cite{fanliu2023book}. Beyond its efficient utilization of spectral and energy resources, the combination of both functionalities offers coordination gains benefited from the joint design and enhances the system performance further\cite{fanliu2022suvey}. In addition, due to the overlapping hardware requirements and signal processing principles, implementing sensing capabilities in existing communication systems such as WLAN\cite{wifisensingoverview} is feasible and convenient while often not necessitating an overhaul or update of the current infrastructure. 
In the realm of robotics, ISAC allows robots to not only interact with their environment more effectively but also to communicate seamlessly within a network. Furthermore, by combining ISAC with traditional sensors like cameras and LiDAR, robots gain a more holistic understanding of their surroundings, leading to more precise and effective actions. 

As another emerging technology in 6G, digital twin (DT) technology is reshaping the landscape of robotics and wireless communication systems\cite{dtfor6g}. Traditionally in robotics, DTs have been used primarily for simulation and modeling purposes. They serve as virtual replicas of robotic systems, providing a platform for testing, analyzing, and optimizing robot designs and behaviors before physical deployment. These approaches have been recently utilized to facilitate the pre-deployment validation and testing of wireless systems\cite{dtforwireless,alkhateeb2023real} and are shown to significantly reduce the expenses and effort traditionally involved in these processes\cite{jiang2023digital}. However, accurately modeling the intricate interactions between the robot and its environment, especially when incorporating ISAC capability into multi-sensor platforms, is complex. Maintaining real-time synchronization between the physical robot and its DT also requires continuously updating and adapting to changes in the robot’s environment and operational parameters. Therefore, the current academic research and practical deployment of real-time DTs for robotics with ISAC integration are still in the early stages. 
On the other hand, addressing and solving these challenges enables a wide array of new applications. 
An accurate DT of the environment could help the agents offload low-level tasks, such as localization. These can then in turn focus their computational resources on solving high-level tasks, such as scene understanding. This aspect coupled with both local and global information also allows for greater energy efficiency and better network planning. 
Toward this end, we developed a hardware-software framework to address the challenges and enable the applications mentioned above. This framework, based on the Robot Operating System (ROS), integrates traditional robotic functionalities of real-time control, perception, and cooperation, along with advanced ISAC system simulation and modeling based on ray-tracing technology. In this paper, we place a particular emphasis on testing its capabilities in indoor localization and navigation, with our contributions being as follows:
\begin{itemize}
    \item We have implemented, to our knowledge, the first DT for ISAC and robotics. This implementation is unique not only in its simulation capabilities but also in offering a real-world interface to actual sensors.
    \item Our DT leverages powerful robotics and ray-tracing technologies and is capable of fusing data from both synthetic and real sources. Additionally, it provides reactive updates triggered by changes in the environment.
    \item We designed our DT to be compatible with off-the-shelf commercial components, which facilitates rapid prototyping and research, thus making it more accessible for studies exploring the intersection of ISAC and robotics.
    \item Our measurement results demonstrate that the localization accuracy provided by the DT, when compared to internal odometry data from the robot, is highly reliable. These results indicate the practical utility of our DT in real-world scenarios.
\end{itemize}

\section{High-Level Overview} \label{sec:overview}
\begin{figure}
    \centering
    \includegraphics[width=\columnwidth, height=0.4\textheight]{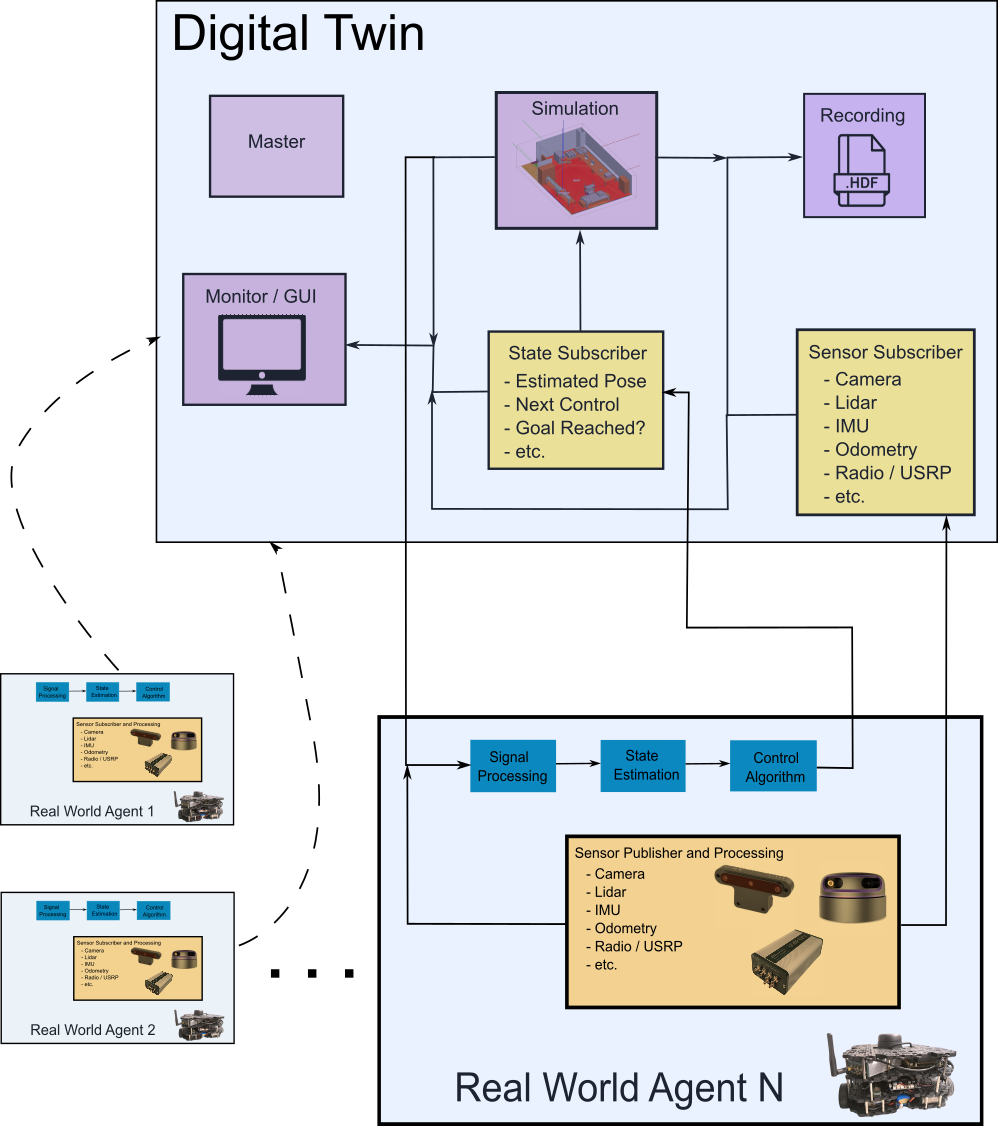}
    \caption{High-level overview of the system architecture}
    \label{fig:dt_high_level}
\end{figure}

The proposed framework consists of a central DT instance running on a remote computer, and multiple autonomous robots as depicted in Figure \ref{fig:dt_high_level}, following a publisher/subscriber model \cite{eugster2003publish} implemented using ROS \cite{quigley2009ros}. 
The DT consists of the following components:

    \textbf{Master Node:} 
    This module provides naming and registration services to the rest of the robots and tracks publishers and subscribers to different topics.
    
    \textbf{Simulation:} 
    This module takes as input a 3D model of the environment, as well as a network configuration that specifies the connectivity between robots in the communication and sensing scenario. Utilizing this input, the module conducts an extensive ray-tracing simulation across all transmit/receive (Tx/Rx) pairs, alongside the modeling of the robots' kinematics and their interactive dynamics. This simulation is updated in real-time by the state of the robots in the real world, and after each simulation timestep, the simulated physical parameters are published.
    
    \textbf{State Subscriber:} 
    The state data published by the robots via WLAN is processed in this module and passed onto the simulation. Examples of state data are the computed position, the next control command, and whether a waypoint on a predefined path has been reached or not.
    
    \textbf{Sensor Subscribers:} 
    Here the sensor data from the robots sent via WLAN is being acquired, processed, and passed onto the nodes responsible for recording and monitoring. Currently supported modalities are camera, LiDAR, odometry, inertial measurement unit (IMU), and data extracted from radio frequency (RF) signals such as channel state information (CSI).
    
    \textbf{Recording and Monitoring:} 
    These nodes aggregate sensor, state, and simulation data, visualize them using the graphical user interface (GUI) and optionally save them to files for subsequent offline processing, including design and verification of advanced algorithms like machine learning (ML).
    
Each of the robots internally implements the following nodes:

    \textbf{RF Signal Processing:} 
    At this stage, the ISAC Tx/Rx processing chains are implemented with the input data generated either by the simulation and sent via WLAN or with the real waveforms transmitted by the other parties on e.g. other frequencies.
    
    \textbf{Sensor Publishers:} 
    Here, the sensor data is being acquired and sent to the other nodes for processing. The robots are currently equipped with cameras, LiDAR, and IMU sensors, but can be very easily extended to other modalities.  
    
    \textbf{State Estimation and Control:} 
    This stage uses data from both simulation and sensors to estimate the current state of the robot such as pose, orientation, or distance to the goal, and compute the next control.
    
\begin{figure}
    \centering
    \includegraphics[width=\columnwidth]{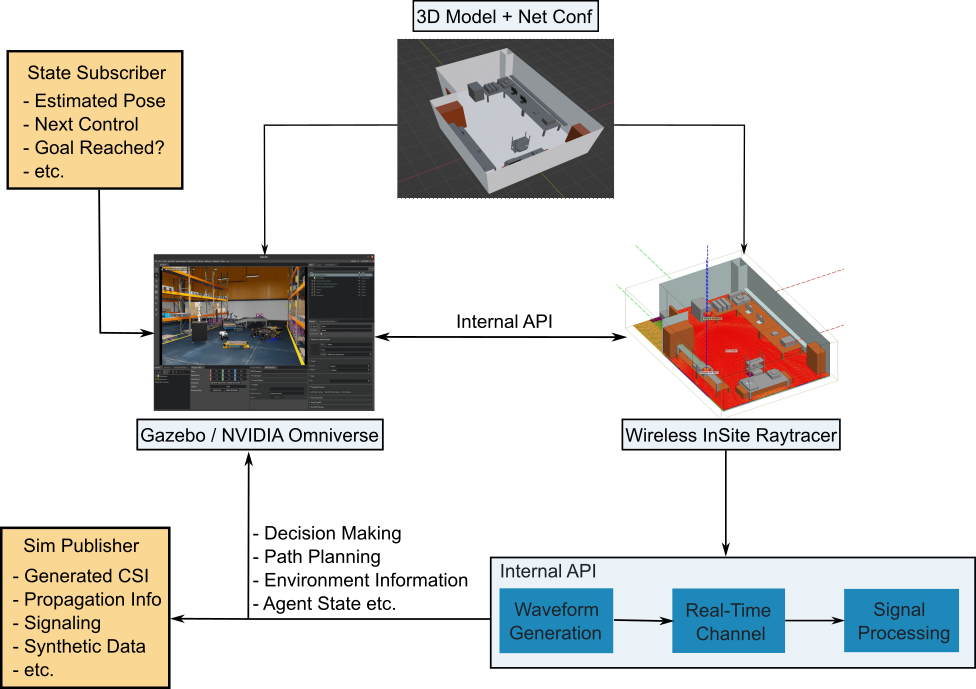}
    \caption{Schematic flow of the simulation}
    \label{fig:sim_overview}
\end{figure}

As depicted in \ref{fig:sim_overview},
 the simulation takes as input a 3D model of the environment with material properties, as well as a network configuration that specifies the Tx/Rx pairs and the initial positions and orientations of the robots. Using this network configuration, the initial propagation parameters are computed by a ray-tracing engine, and the communication links between the Tx/Rx pairs are simulated on the physical layer, where MIMO-OFDM signaling is employed. Afterward, the observations for each of the robots are generated by the simulator, followed by the computation of the control command and state update. To accurately model and visualize the kinematics of each robot, we support the integration with the Gazebo \cite{koenig2004gazebo} and NVIDIA Omniverse \cite{nvidia2023omniverse} simulation engines.

\section{Implementation Aspects} \label{sec:impl}
In this section, we elaborate upon the implementation of the simulation, the construction of the 3D model and on the realization of the autonomous agents using the Turtlebot3 \cite{turtlebot2023website} robotic platform.
\subsection{Communications, Sensing and Robotics Modelling}
We consider a network $\mathcal{G} = \left(\mathcal{A}, \mathcal{E}\right)$ of $Q$ transmitters and $V$ receivers, modelled as a undirected graph structure without self-loops, with the collection of all nodes $\mathcal{A}=\{\mathcal{Q},\mathcal{V}\}$ consisting of the set of transmitters $\userset=\{1, \dots, Q\}$ and the set of receivers $\rxset=\{1, \dots, V\}$, and $\mathcal{E}$ being the set of edges between nodes.

The wireless link is modelled in each simulation step as follows. Transmitter $q\in\userset$ sends signal $\x_{qnk}$ by beamforming the scalar communication and/or sensing symbol $d_{qnk}$ at subcarrier $n \in \carrierset_{q}$ and time instance $k \in \symset_{q}$ with the beamformer $\wa_{qnk} \in \C^{N_{T_q}}$ and scaling it with the power factor $p_{qnk}$.
$\symset_{q}$ and $\carrierset_{q}$ denote the symbol and subcarrier sets allocated to each user $q$. 
Assuming a total of $\nsc$ subcarriers and $K$ available symbols for resource allocation, the 
resource sets  $\resourceset_{q} = \carrierset_q \times \symset_q$ of each user satisfy $\cup_{q=1}^{Q} \carrierset_{q} \subseteq \carrierset = \{1, \dots, \nsc\}$ and  
$\cup_{q=1}^{Q} \symset_{q} \subseteq \symset = \{1, \dots, K\}$,
with the total power available to all resource elements being $P_{q}$, i.e. $\sum_{(n, k)\in\resourceset_q}\normsq{\x_{qnk}}\leq P_{q}$.
The resulting signals propagate through the legitimate channels $\ch_{vqnk}\in\mathbb{C}^{\nrv\times \ntq}$, are corrupted by white noise $\z_{nk}\sim\gauss{\cov{\z_{nk}}}$ and then finally arrive at the $\nrv$ receive antennas of receiver $v\in\rxset$.
More formally:
\begin{align}\label{eq:sys_model}
    &\x_{qnk} = \alpha_{qnk}\sqrt{p_{qnk}}\wa_{qnk} d_{qnk} \in \C^{\ntq},\\ \label{eq:rx_signal}
    &\y_{vnk} = \sum_{q\in\mathcal{E}_{v}}\ch_{vqnk}\x_{qnk} + \z_{nk} \in \C^{\nr}
\end{align}
Here, $\alpha_{qnk} = 1$ if user $q$ occupies resource element $(n, k)$  and $0$ else, and $\mathcal{E}_{v}$ is the set of incoming edges of receiver $v$. 
The MIMO-OFDM propagation channels $\ch_{qnk}$ are modelled as beam-space channels, i.e.
\begin{align}\nonumber
    \ch_{vqnk} &= \sum_{l=1}^{L_{vq}} 
    b_{H_{vq}, l} \ex^{\jma 2\pi\omega_{nk}(\nu_{vq,l}, \tau_{vq,l})} \cdot \\ 
    &\cdot \bm{a}_{\nrv}(\btheta_{vq,l}) \bm{a}_{\ntq}\tran(\bpsi_{vq, l}) \label{eq:beamspace}
\end{align}
with $L_{\cdot}$ being the number of resolvable paths for each channel, $\bm{a}_{\cdot}(\btheta)$ denote the steering vectors at each terminal,
and $\btheta_{\cdot, l}$, $\bpsi_{\cdot, l}$, $b_{\cdot, l}$ the azimuth and elevation directions of arrival, directions of departure and path gain,
corresponding to the $l$-th resolvable path, respectively. The term $\omega_{nk} = k \nu T_{s} - n \tau\Delta f$ quantifies the phase shift caused by the Doppler shifts $\nu_{\cdot,l}$ and propagation delays $\tau_{\cdot, l}$ of each multipath component (MPC) $l$.
The parameters 
\begin{equation}
    \bxi_{vq} = [\bm{b}_{vq}\tran, \btau_{vq}\tran, \bm{\nu}_{vq}\tran, \btheta_{vq}\tran, \bpsi_{vq}\tran]\tran \in \C^{5L_{vq}}
\end{equation}
of the beamspace channel are computed by the ray-tracing engine in each simulation step depending on the state vector of each agent. We use the commercially available Remcom WirelessInSite \cite{remcom2023wirelessinsite} raytracer, but provide a unified software interface which can enable seamless integration with NVIDIA Sionna \cite{hoydis2022sionna} and MaxRay \cite{arnold2022maxray}. Note that the raytracer output also depends on the scenario geometry and material properties of the object present in the room. These are initialized a-priori using a 3D model of the environment with material information, which we elaborate upon in Section~\ref{sec:3d_model}.

We now describe the state-action space representation of each agent, which governs both their decision-making and control computation, as well as the raytracer output. The states of agent $\bm{s}_{a, t}, a\in\mathcal{A}$ at each timestep $t$ include but not restricted to its pose information $[x_{a,t}, y_{a,t}, z_{a,t}, \alpha_{a,t}, \beta_{a,t}, \gamma_{a, t}]\in\R^{6}$. Here $x,y,z$ denote the position in the Carthesian plane and $\alpha, \beta, \gamma$ the Euler angles. Other variables such as current beamforming vector and power allocation might also be included. 
The observations $\obs_{a,t}\in \R^{O_{a}}$ at timestep $t$ stem from the measurements $\bm{m}_{a,t}\in\C^{M_{a}}$ made by the agent, which directly depend on the received MIMO-OFDM signals and possibly other variables, such as raytracer output $\{\bxi_{aq}\}_{q\in\userset}$, as well as on the current state. Here $O_{a}$ and $M_{a}$ denote the dimensionalities of the observation and measurement vector for each agent $a$.
Thus the state-space representation of each agent is governed by following non-linear, stochastic set of difference equations:
\begin{align}\label{eq:state}
    \bm{s}_{a, t} &= \bm{f}(\bm{s}_{a, t-1}, \bm{u}_{a, t-1}) + \bm{\varepsilon_{s_{t}}}\\ \label{eq:obs}
    \obs_{a, t} &= \bm{g}(\bm{s}_{a, t}, \bm{m}_{a,t}) + \bm{\varepsilon_{\obs_{t}}}
\end{align}
where $\bm{f}, \bm{g}$ are possibly non-linear functions, $\bm{\varepsilon_{s_{t}}}, \bm{\varepsilon_{\obs_{t}}}$ are stochastic noise terms, usually chosen as zero-mean Gaussians with constant variance, and $\bm{u}_{a,t}$ is the control input/action with dimensionality $U_{a}$. 
Note, that the graph being undirected means that we also allow the receiver nodes to be transmitting, thus enabling the simulation of duplex communication and sensing links. Furthermore, the fact that self-cycles are not allowed might pose a problem for monostatic ISAC applications. On the other hand, this can be easily solved by instantiating dummy transmitter nodes at the positions of the objects of interest, which just implement the scattering characteristics of interest.
At last, Algorithm \ref{alg:sim_step} summarized the simulation procedure in pseudo-code.
\begin{algorithm}[t]
    \caption{Simulation Overview.}
    \label{alg:sim_step}
    \textbf{Input} Initial states (poses) $\bm{s}_{a, 0}$ and control actions $\bm{u}_{a, 0}$ of the agents, network graph $\mathcal{G}$ and 3D model of the environment.
    \vspace{1mm} \hrule \vspace{1mm}
    \begin{algorithmic}[1]
    \While{not terminated}
        \State Update agent states using the past control command and state using Equation \eqref{eq:state}.
        \State Update propagation parameters based on the new agent poses by calling the raytracer.
        \State Generate the MIMO-OFDM signals using \eqref{eq:rx_signal} and other relevant data for each agent.
        \State Compute measurements and observations for each agent. 
        \State Estimate new state for each agent using equation \eqref{eq:obs}.
        \State Generate the next control command for each agent.
    \EndWhile
    \end{algorithmic}
\end{algorithm}

Furthermore, the state evolution in equation \eqref{eq:state} can either be implemented by hand or be the output of robotics simulation tools, in our case NVIDIA Omniverse or Gazebo. Notice that, while the pseudo-code in Algorithm \ref{alg:sim_step} subtly implies that the updates are being done sequentially, this might not be the case and further synchronization between the data for each agent must be ensured. 
At last, if one wishes to simulate a scenario involving static transmitters and dynamic receivers, the ray-tracing output can be pre-computed for all possible positions of the dynamic receivers and stored in a database. This is valid since the rays only need to be generated once, which is the compute-intensive step in ray-tracing methods.    
\subsection{3D Model Creation}\label{sec:3d_model}
In the following, we elaborate upon the creation of the 3D model of the experimental room at the Advanced Communications and Embedded Security (ACES) Lab of the TU Munich. At first, a ground plan of the room and a list of all objects inside was created. We only included bigger objects such as switch boxes, measurement devices, tables, and chairs. Afterwards, the length, width and height of each objects, were measured and the surface material was determined either by inspection or by the corresponding entries in the data sheets.

\begin{figure}
    \centering
        \includegraphics[width=\columnwidth, height=0.25\textheight]{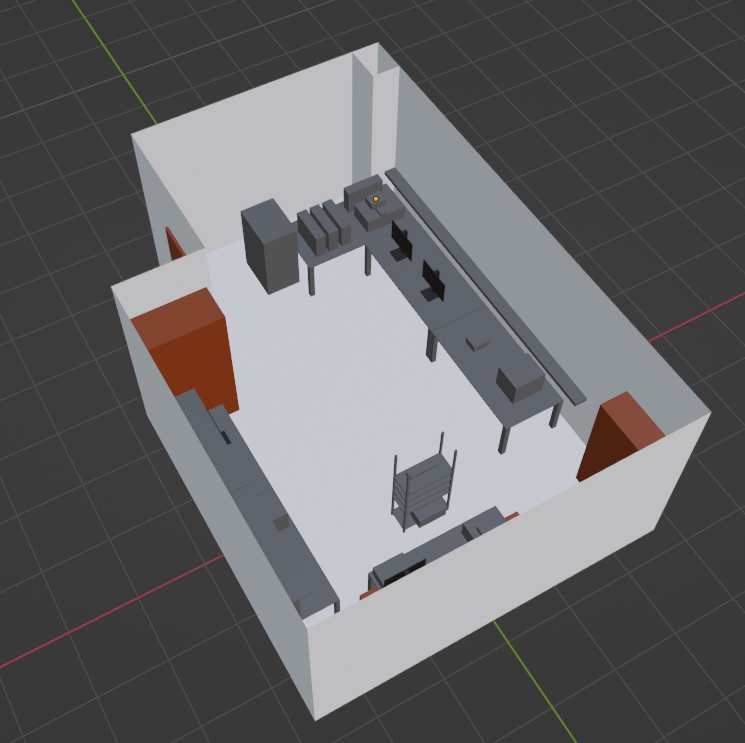}
        \caption{Blender rendering of the experimental space}
        \label{fig:aces_blender}
\end{figure}

Then the objects and the ground plan were recreated in Blender, and the resulting 3D model was exported into DAE, STL and USD formats. Finally, this model alongside material lists were saved to disk, in order to be imported into the ray-tracing and robotic simulation software later. 
Figure \ref{fig:aces_blender} shows the rendered view. We note that with this manual approach, the scale in the 3D model is the same as in reality. On the other hand, it might incur errors regarding the relative positioning of the objects to each other. These deviations are not straightforward to measure and their impact depends on the task at hand. We will elaborate in Section \ref{sec:case_study} upon this and show how the modeling error can lead to errors in localization performance. Nonetheless, this could be improved by e.g. 3D LiDAR scans of the room, which we leave for future work.
\subsection{Robotic Platform Implementation}\label{sec:robot}
\begin{figure}
    \centering
    \includegraphics[width=0.8\columnwidth]{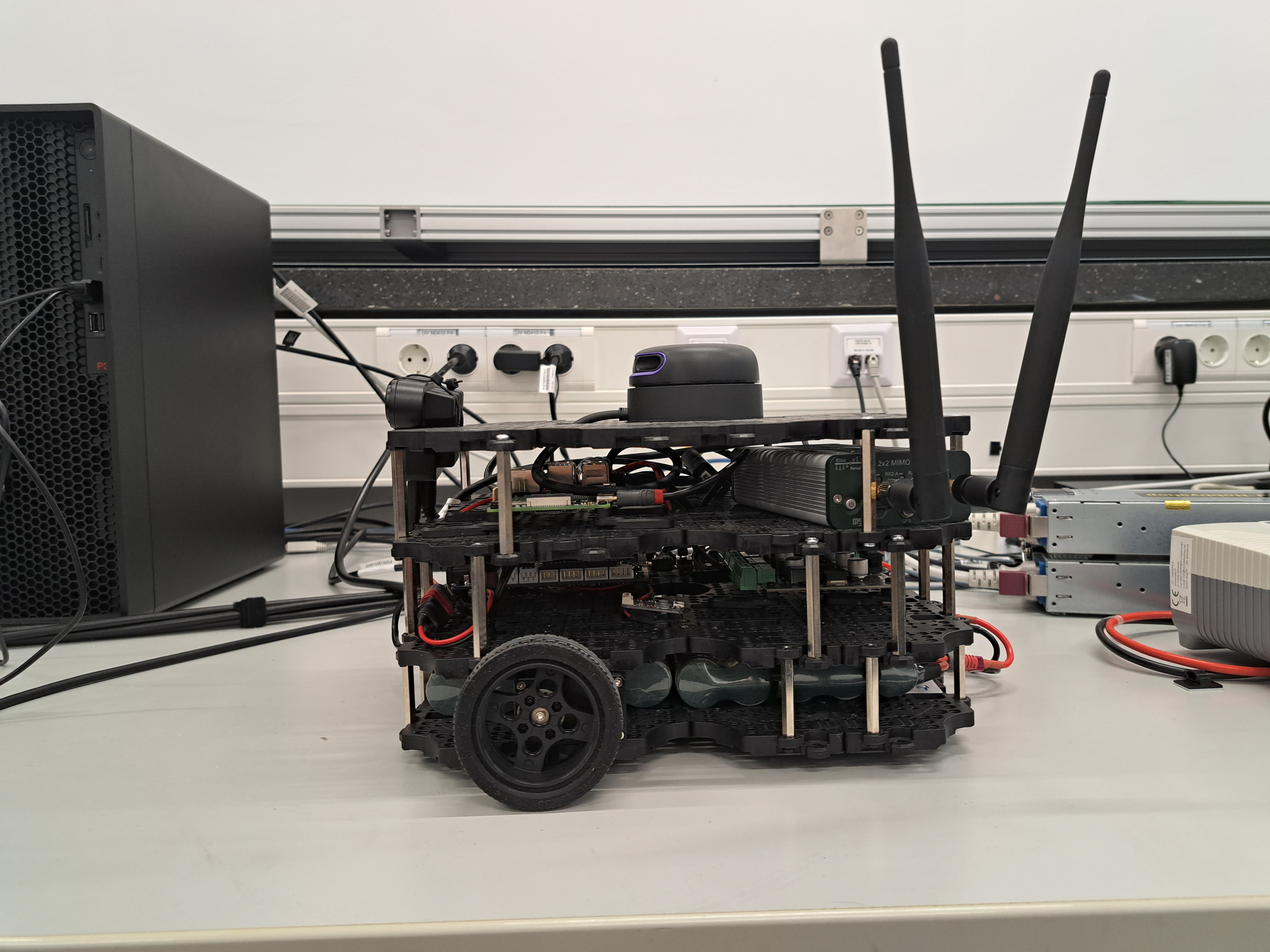}
    \caption{Modified Turtlebot3 platform used in experiments}
    \label{fig:robot}
\end{figure}

The robotic platform used in Section \ref{sec:case_study} is a modified version of the Turtlebot3 robot, depicted in Figure \ref{fig:robot}. The Turtlebot3 is a small, affordable, programmable, ROS-based mobile robot, which can be easily customized and extended with multiple sensors and compute platforms. 

On the bottom layer of the robot, one \SI{12}{\volt} battery and one \SI{24}{\volt} battery are mounted alongside the motors. The former battery pack is used to power the OpenCR motor controller on the upper layer and the latter connects to a InnoMaker LM2596 4-channel DC to DC buck converter. The motor controller comes with 3-axis gyroscope, 3-axis accelerometer, and digital motion processor, and is used to actuate the wheels by the commands of the main compute unit. The converter splits the input voltage in 4 channels which can be configured to deliver \SI{12}{\volt}, \SI{5}{\volt}, \SI{3.3}{\volt} and adjustable voltages. Each output channel can power up to 2 devices simultaneously. Furthermore, the robot is equipped with two compute units: one central computer (Raspberry Pi 4) which aggregates sensor and DT data and computes the controls, and one NVIDIA Jetson Xavier embedded GPU used to offload some of the more intensive computations. The main computing unit is connected to a Luxonis OAK-D stereo camera with object detection capabilities and to a Slamtec LiDAR sensor by USB. A 4-port Renkforce USB switch is used to interconnect the main compute unit, the embedded GPU, and the software-defined radio (Ettus USRP E312), which is used to acquire CSI data and solve other communication tasks. 
\section{Case Study: Joint Communications and Localization} \label{sec:case_study}
In this section, we investigate how our DT can be used to make real-time decisions in the physical world, using fingerprinting(FP)-based indoor localization \cite{zayets2017robust} and navigation as an example, as well asses the communications modeling capabilities. 
FP-based localization methods consist of two stages, namely training and deployment. In the training stage, the fingerprints of multiple access points (APs) are measured at known positions in the environment and stored in a database. In the deployment stage, a position estimate is obtained by matching the newly measured fingerprints to the ones in the database. 
In this paper, we adopt the multipath component analysis (MCA) algorithm \cite{zayets2017robust}, which employs the multipath delay profiles (MDPs) as the fingerprints. 

Constructing the database is an expensive process, especially when the room environment or the objects change frequently. Thus, the first goal of this study is to investigate the feasibility of using a DT in the training phase of MCA to reduce the effort of creating databases in the real world. This also implies investigating the fidelity of the DT concerning positions and distances in the real world.
 The second goal is to have an initial, qualitative assessment of the communications simulation.
\subsection{Experimental Setup}\label{sec:setup}
\begin{figure}[htbp]
    \centering
    \includegraphics[width=0.8\columnwidth, height=0.25\textheight]{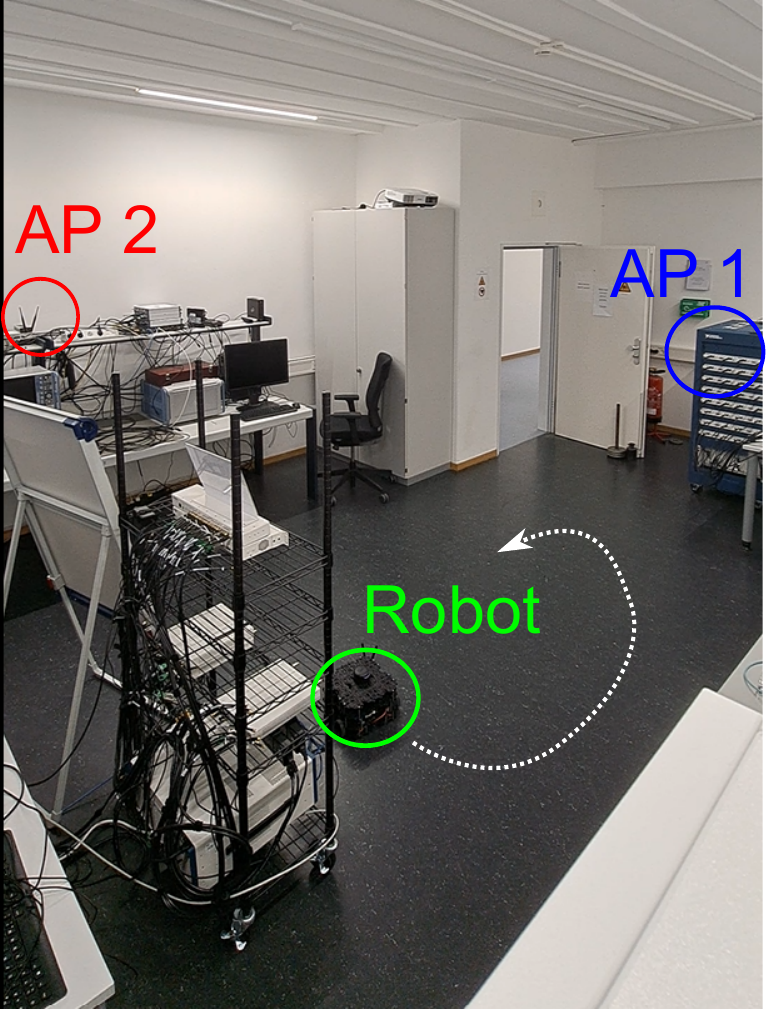}
    \caption{View of the experimental setup. The white curve is a qualitative representation of the circular trajectory the robot needs to follow.}
    \label{fig:exp_setup}
\end{figure}

The setup modeled and simulated in the DT consists of two APs and a single robot as depicted in Figure \ref{fig:exp_setup}. 
The first AP is a MIMO station with $32$ antennas and the other one is a single-antenna radio unit, both operating at $\SI{2.4 }{\giga\hertz}$. The robot itself is equipped with two antennas, as depicted in Figure \ref{fig:robot}, and all parties communicate using OFDM signals with $N=1024$ subcarriers of spacing $\Delta f = \SI{78.125}{\kilo\hertz}$. The two APs convey pilots and communication data to the robot, which needs to localize itself and navigate a predefined path. 
\subsection{Measurement Results and Discussion}
To achieve the first goal of this study, the raytracer output is computed using the network configuration mentioned in Section \ref{sec:setup} for each point on the floor in a $\SI{5}{\centi\meter}$ spacing. Afterward, the obtained database is loaded into the robot's memory. In the deployment phase, the robot sends its position estimate to the DT, which then computes the MDP at the robot's position. Note that this fingerprint is generally not contained in the database. Finally, the DT sends back the computed fingerprint, which the robot then uses for its next position estimate. 
Note that while the robot software supports MDP estimation using pilot data, we explicitly use the raytracer output to further stress-test our implementation.

\begin{figure}
    \centering
    \includegraphics[width=\columnwidth]{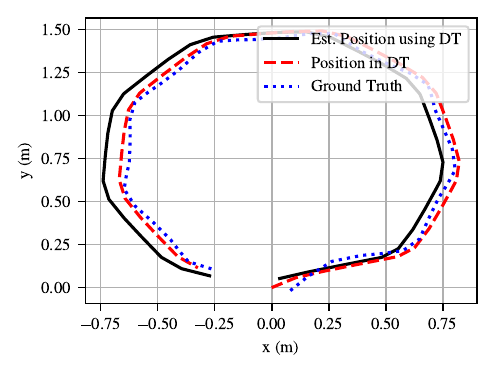}
    \caption{Robot trajectories}
    \label{fig:traj}
\end{figure}

Figure \ref{fig:traj} shows three robot trajectories, namely the trajectory estimated by the robot using data from the DT, the trajectory of the robot in the DT, and the trajectory reconstructed by odometry data computed using the robot's IMU, which we use as the ground truth.
The trajectories are shown to be close to each other, which indicates that both the positioning algorithm and the mapping in the DT perform well qualitatively. 

\begin{figure}
    \centering
    \includegraphics[width=\columnwidth]{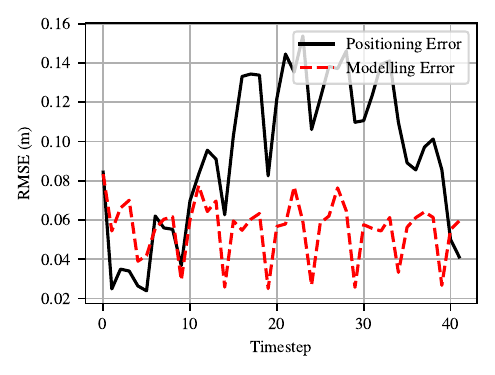}
    \caption{Positioning and modelling errors incurred by MCA and DT}
    \label{fig:errs}
\end{figure}

 In order to support this argument, we plot the positioning and modeling error in Figure \ref{fig:errs}. The positioning error is computed by means of the RMSE between the robot estimates and the ground truth, and the modeling error is the RMSE between the position in the simulation and the ground truth. We can observe that the maximum positioning error is about $\SI{16}{\centi\meter}$, which is in concordance with the simulation results in \cite{zayets2017robust} and also fulfills the requirements for indoor positioning outlined in \cite{italiano2023tutorial}. Note, that this error is not only influenced by the algorithm performance but also by the modeling error, also depicted in Figure \ref{fig:errs}. This error mainly stems from the fact that the initial point used for odometry calibration can not be mapped perfectly in the DT by our manual approach. Nonetheless, the error incurred is quite low, never exceeding $\SI{8}{\centi\meter}$. A possible solution for this would be the use of a visual and inertial odometry system to initialize the positions of the objects.
 
\begin{figure}
    \centering
    \includegraphics[width=\columnwidth]{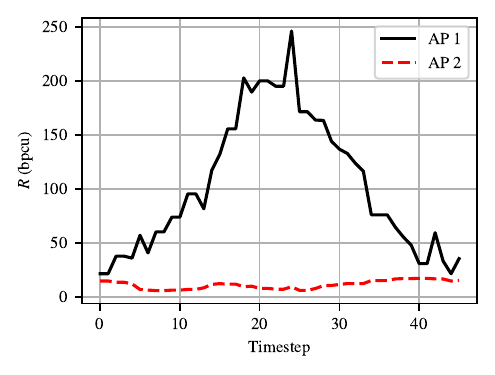}
    \caption{Acheivable rates simulated by the DT}
    \label{fig:rates}
\end{figure}

 At last, Figure \ref{fig:rates} shows the achievable rates of both communication links as simulated by our framework over $K = 14$ OFDM symbols in each simulation timestep. We observe that the magnitudes differ drastically, which is to be expected since AP 1 employs a massive MIMO setup for communication, while AP 2 only possesses a single antenna. Furthermore, we observe that the achievable rates increase with decreasing distance from the respective transmitter when the robot moves along the curve represented qualitatively in Figure \ref{fig:exp_setup}, which can be preeminently seen in the case of AP 1. This furthermore confirms the validity of our DT implementation.  

\section{Conclusions and Future Work} \label{sec:conclusions}
We have proposed, designed, and implemented a DT platform for ISAC and robotics, which not only leverages powerful robotics and ray-tracing technologies for simulation but also enables seamless integration with real-world sensors. Experimental measurement results show our DT can be used to both achieve good indoor navigation performance and emulate communication links. Future work will be dedicated to further improving the fidelity of our DT by incorporating 3D-LiDAR scans, as well as visual and inertial odometry, and to the validation of communication fidelity by experimental measurements.
\bibliography{IEEEabrv, 2023JCAS_dt4isac}

\begin{thebibliography}{10}
\providecommand{\url}[1]{#1}
\csname url@samestyle\endcsname
\providecommand{\newblock}{\relax}
\providecommand{\bibinfo}[2]{#2}
\providecommand{\BIBentrySTDinterwordspacing}{\spaceskip=0pt\relax}
\providecommand{\BIBentryALTinterwordstretchfactor}{4}
\providecommand{\BIBentryALTinterwordspacing}{\spaceskip=\fontdimen2\font plus
\BIBentryALTinterwordstretchfactor\fontdimen3\font minus \fontdimen4\font\relax}
\providecommand{\BIBforeignlanguage}[2]{{%
\expandafter\ifx\csname l@#1\endcsname\relax
\typeout{** WARNING: IEEEtran.bst: No hyphenation pattern has been}%
\typeout{** loaded for the language `#1'. Using the pattern for}%
\typeout{** the default language instead.}%
\else
\language=\csname l@#1\endcsname
\fi
#2}}
\providecommand{\BIBdecl}{\relax}
\BIBdecl

\bibitem{walid6g}
W.~Saad, M.~Bennis, and M.~Chen, ``A vision of {6G} wireless systems: Applications, trends, technologies, and open research problems,'' \emph{IEEE Network}, vol.~34, no.~3, pp. 134--142, 2020.

\bibitem{holger6g}
P.~Schwenteck, G.~T. Nguyen, H.~Boche, W.~Kellerer, and F.~H.~P. Fitzek, ``{6G} perspective of mobile network operators, manufacturers, and verticals,'' \emph{IEEE Networking Letters}, vol.~5, no.~3, pp. 169--172, 2023.

\bibitem{fanliu2023book}
F.~Liu, C.~Masouros, and Y.~C. Eldar, \emph{Integrated Sensing and Communications}.\hskip 1em plus 0.5em minus 0.4em\relax Springer Nature, 2023.

\bibitem{fanliu2022suvey}
F.~Liu, Y.~Cui, C.~Masouros, J.~Xu, T.~X. Han, Y.~C. Eldar, and S.~Buzzi, ``Integrated sensing and communications: Toward dual-functional wireless networks for {6G} and beyond,'' \emph{IEEE Journal on Selected Areas in Communications}, vol.~40, no.~6, pp. 1728--1767, 2022.

\bibitem{wifisensingoverview}
R.~Du, H.~Hua, H.~Xie, X.~Song, Z.~Lyu, M.~Hu, Y.~Xin, S.~McCann, M.~Montemurro, T.~X. Han \emph{et~al.}, ``An overview on ieee 802.11 bf: {WLAN} sensing,'' \emph{arXiv preprint arXiv:2310.17661}, 2023.

\bibitem{dtfor6g}
A.~Masaracchia, V.~Sharma, B.~Canberk, O.~A. Dobre, and T.~Q. Duong, ``Digital twin for {6G}: Taxonomy, research challenges, and the road ahead,'' \emph{IEEE Open Journal of the Communications Society}, vol.~3, pp. 2137--2150, 2022.

\bibitem{dtforwireless}
L.~U. Khan, Z.~Han, W.~Saad, E.~Hossain, M.~Guizani, and C.~S. Hong, ``Digital twin of wireless systems: Overview, taxonomy, challenges, and opportunities,'' \emph{IEEE Communications Surveys \& Tutorials}, vol.~24, no.~4, pp. 2230--2254, 2022.

\bibitem{alkhateeb2023real}
A.~Alkhateeb, S.~Jiang, and G.~Charan, ``Real-time digital twins: Vision and research directions for {6G} and beyond,'' \emph{IEEE Communications Magazine}, 2023.

\bibitem{jiang2023digital}
S.~Jiang and A.~Alkhateeb, ``Digital twin based beam prediction: Can we train in the digital world and deploy in reality?'' \emph{arXiv preprint arXiv:2301.07682}, 2023.

\bibitem{eugster2003publish}
P.~T. Eugster, P.~A. Felber, R.~Guerraoui, and A.-M. Kermarrec, ``{The Many Faces of Publish/Subscribe},'' \emph{ACM Comput. Surv.}, vol.~35, no.~2, p. 114–131, jun 2003.

\bibitem{quigley2009ros}
M.~Quigley, K.~Conley, B.~Gerkey, J.~Faust, T.~Foote, J.~Leibs, R.~Wheeler, A.~Y. Ng \emph{et~al.}, ``{ROS: an open-source Robot Operating System},'' in \emph{ICRA workshop on open source software}, vol.~3, no. 3.2.\hskip 1em plus 0.5em minus 0.4em\relax Kobe, Japan, 2009, p.~5.

\bibitem{koenig2004gazebo}
N.~Koenig and A.~Howard, ``{Design and Use Paradigms for Gazebo, an Open-Source Multi-Robot Simulator},'' in \emph{2004 IEEE/RSJ International Conference on Intelligent Robots and Systems (IROS)}, vol.~3, 2004, pp. 2149--2154 vol.3.

\bibitem{nvidia2023omniverse}
\BIBentryALTinterwordspacing
``{NVIDIA Omniverse Platform}.'' [Online]. Available: \url{https://developer.nvidia.com/omniverse}
\BIBentrySTDinterwordspacing

\bibitem{turtlebot2023website}
\BIBentryALTinterwordspacing
``Turtlebot3.'' [Online]. Available: \url{https://emanual.robotis.com/docs/en/platform/turtlebot3/overview/}
\BIBentrySTDinterwordspacing

\bibitem{remcom2023wirelessinsite}
\BIBentryALTinterwordspacing
``{Wireless InSite, Remcom Inc.}'' [Online]. Available: \url{https://www.remcom.com/wireless-insite-em-propagation-software}
\BIBentrySTDinterwordspacing

\bibitem{hoydis2022sionna}
J.~Hoydis, S.~Cammerer, F.~{Ait Aoudia}, A.~Vem, N.~Binder, G.~Marcus, and A.~Keller, ``{Sionna: An Open-Source Library for Next-Generation Physical Layer Research},'' \emph{arXiv preprint}, Mar. 2022.

\bibitem{arnold2022maxray}
M.~Arnold, M.~Bauhofer, S.~Mandelli, M.~Henninger, F.~Schaich, T.~Wild, and S.~ten Brink, ``{MaxRay: A Raytracing-based Integrated Sensing and Communication Framework},'' in \emph{2022 2nd IEEE International Symposium on Joint Communications \& Sensing (JC\&S)}, 2022, pp. 1--7.

\bibitem{zayets2017robust}
A.~Zayets and E.~Steinbach, ``Robust wifi-based indoor localization using multipath component analysis,'' in \emph{2017 International Conference on Indoor Positioning and Indoor Navigation (IPIN)}, 2017, pp. 1--8.

\bibitem{italiano2023tutorial}
L.~Italiano, B.~C. Tedeschini, M.~Brambilla, H.~Huang, M.~Nicoli, and H.~Wymeersch, ``A tutorial on {5G} positioning,'' 2023.

\end{thebibliography}
\bibliographystyle{IEEEtran}
\end{document}